\documentclass{article}

\usepackage{microtype}
\usepackage{graphicx}
\usepackage{subfigure}
\usepackage{booktabs} 
\usepackage{multirow}
\usepackage{amssymb}
\usepackage{amsmath}
\usepackage{float} 
\usepackage{scalerel} 

\DeclareMathOperator{\sign}{sign}

\usepackage[accepted]{icmlw2019generalization}

\icmltitlerunning{Towards Task and Architecture-Independent Generalization Gap Predictors}

\begin{document}

\twocolumn[
\icmltitle{Towards Task and Architecture-Independent Generalization Gap Predictors}



\icmlsetsymbol{equal}{*}

\begin{icmlauthorlist}
\icmlauthor{Scott Yak}{goo}
\icmlauthor{Javier Gonzalvo}{goo}
\icmlauthor{Hanna Mazzawi}{goo}
\end{icmlauthorlist}

\icmlaffiliation{goo}{Google Research, NYC, USA}

\icmlcorrespondingauthor{Javier Gonzalvo}{xavigonzalvo@google.com}

\icmlkeywords{Machine Learning, Generalization Gap, Excess Risk, Meta-learning ICML}

\vskip 0.3in
]



\printAffiliationsAndNotice{}  

\begin{abstract}
Can we use deep learning to predict when deep learning works? Our results suggest the affirmative. We created a dataset by training 13,500 neural networks with different architectures, on different variations of spiral datasets, and using different optimization parameters. We used this dataset to train task-independent and architecture-independent generalization gap predictors for those neural networks.
We extend \citet{jiang2018predicting} to also use DNNs and RNNs and show that they outperform the linear model, obtaining $R^2=0.965$. We also show results for architecture-independent, task-independent, and out-of-distribution generalization gap prediction tasks. Both DNNs and RNNs consistently and significantly outperform linear models, with RNNs obtaining $R^2=0.584$.
\end{abstract}

\section{Introduction}
\label{introduction}
\textit{Generalization} measures the ability of a model to predict the labels correctly on previously unseen data points. The difference between model performance on training data versus unseen data is known as the {\it generalization gap}.
Despite generalization having been intensively studied in the past few decades, understanding when a DNN is likely to generalize with high confidence is still an open problem.

The recent work of \citet{zhangBHRV16} has adequately summarized the main problems with DNN generalization: understanding the effect of model complexity, overfitting and the role of regularization. On one hand, it was shown that the number of parameters is not correlated with model overfitting, suggesting that parameter counting cannot indicate the true complexity of deep neural networks. This has to do with the fact that DNNs generalize even when the models are clearly overparameterized~\cite{allenzhu18overparams}.
On the other hand, even though DNNs can fit noise, the absence of all regularization does not necessarily imply poor generalization even in the cases of tighter bounds~\cite{bartlettFT17}. This is because traditional regularization techniques can help fine tuning a model to avoid overfitting but they cannot guarantee good model generalization~\cite{zhangBHRV16}.

The traditional approach to improve generalization is to alleviate overfitting via measures like the Rademacher complexity or VC dimension and their corresponding generalization gap bounds. However, these approaches suffer due to the very high complexity of the hypothesis class when DNNs are very deep \cite{bartlettFT17}. Even though numerous improvements have been presented \cite{neyshaburBMS17, arora18}, these approaches still seem insufficient to describe the behavior of overparametrized DNNs \cite{neyshabur18overparametrized}.

Given that traditional strategies to deal with generalization do not seem to understand it well, a new area of research emerged in order to try to predict the generalization gap as opposed to improving its bounds. The idea is to extract features from the neural network during training time and use them to predict the generalization gap. The main purpose of having an accurate predictor for the generalization gap is that: (1) it eliminates the need of a holdout set; (2) it would facilitate neural architecture search (NAS)~\cite{zophVSL17nasnet}; (3) it might also shed light on how to calculate new tighter complexity measures for the capacity of neural networks; and (4) it can also be used as a type of regularization in loss functions to reduce overfitting.



Most of the recent efforts have focused on finding a set of features that is predictive of the generalization gap \cite{sokolic2017robust, bartlettFT17, elsayed2018large,jiang2018predicting}. The work of \citet{jiang2018predicting} demonstrated how margin signatures in multiple layers of a DNN can predict the generalization gap with high correlation with respect to the actual value. Despite those successes in predicting generalization accurately, state-of-the-art examines predicting generalization on a single task and identical network topologies. While in general this is an obvious limitation, it is especially problematic in the NAS field, where a search algorithm explores a large number of architectures.

While this paper can be seen as a direct extension to \citet{jiang2018predicting}'s work, which introduced the procedure of predicting generalization gap from margin distributions, our contributions are as follows:

\textbf{Simpler feature engineering.}\\
In order to handle different neural network architectures, we present two different approaches. In the first approach, we propose to sum over the layer-specific features to obtain a fixed-dimension vector.  In the second approach, we use an RNN where the recurrence handles input features of an unknown number of layers straightforwardly. Both approaches achieve competitive results. See Section~\ref{sec:features} for more details.

\textbf{Depth-independent vs. depth-agnostic modeling.}\\
Unlike previous work, our approach is not only model-agnostic but it is also model-independent.
We predict the generalization gap across DNNs with different topologies simultaneously on different variations of the spiral datasets with a final coefficient of determination ($R^2$) of 0.965.

\textbf{Task-independent vs. task-agnostic modeling.}\\
To our knowledge, this work is the first attempt to model the generalization gap of DNNs trained on different datasets using the same model.

\textbf{Out-of-distribution vs. out-of-sample evaluation.}\\
Train and test sets are drawn from different data distributions. We train the generalization gap predictors on DNNs trained with a subset of hyperparameters and datasets, then evaluate them on DNNs trained on unseen hyperparameters or datasets.

\textbf{More training examples}\\
Our models are trained with around 400 examples (dataset-dependent case) and with about 8000 examples (dataset-independent case) while previous generalization gap predictors were trained with around 200-300 examples.



\section{Predicting Generalization Gap}
\setlength{\abovedisplayskip}{5pt}
\setlength{\belowdisplayskip}{5pt}
Consider a binary classification problem with dataset $\mathcal{D} \subseteq \mathcal{X} \times \{-1, +1\}$, and consider the class of binary classifiers that are non-linear functions $F = \{ f: \mathcal{X} \rightarrow \mathbb{R} \}$ that are trained on the training set $S \sim \mathcal{D}^m$. Predictions on input $\boldsymbol{x} \in \mathcal{X}$ are obtained by taking the sign of $f(\boldsymbol{x})$. From each classifier $f$, we compute the ground truth label, the generalization gap $g(f)$ (i.e., training accuracy - test accuracy).
We want to find a \emph{generalization gap predictor} ($\hat{g}$ or GGP) that approximates $g$ as closely as possible \emph{without} access to any data beyond the training set, i.e., $\hat{g}(f, S) \approx g(f, S, \mathcal{D})$.

\subsection{Margin Signatures}
As with any other regression problem, we first need to extract features from each binary classifier $f \in F$, and use them as the inputs to the predictor. We restrict $F$ to the class of fully-connected feed-forward DNNs with no skip connections. This restriction allows us to extract from each $f \in F$ a set of \emph{margin signatures} (as defined in \citet{jiang2018predicting}).

Given a DNN $f \in F$, its \emph{decision boundary} is defined as the set of points in $\mathcal{X}$ where $f(\boldsymbol{x}) = 0$.
Along with an input $\boldsymbol{x}$, and label $y$, the signed distance $d^*_{f, \boldsymbol{x}, y}$ is defined as follows:
\begin{equation}
\label{eq:d}
d^*_{f,\boldsymbol{x}, y} \triangleq \sign{(yf(\boldsymbol{x}))} \min_{\mathbf{\delta}} \|\mathbf{\delta}\|_2 \enskip \text{s.t.} \enskip f(\boldsymbol{x} + \mathbf{\delta}) = 0.
\end{equation}

However, as discussed in~\citet{jiang2018predicting}, $d^*_{f, \boldsymbol{x}, y}$ is intractable to compute, so we adopt a similar approximation scheme defined in ~\citet{elsayed2018large}, and instead measure a first-order Taylor approximation to $d^*_{f, \boldsymbol{x}, y}$. Also, generalizing Equation~\ref{eq:d} to multiple hidden layers of a DNN we obtain:
\begin{equation}
\label{eq:approx_dist_general}
    d_{f, \boldsymbol{x}^l, y} = \frac{yf(\boldsymbol{x}^l)}{\|\nabla_{\boldsymbol{x}^l} f(\boldsymbol{x}^l) \|_2},
\end{equation}
where given an input $\boldsymbol{x} \in \mathcal{X}$ to a DNN $f$, $\boldsymbol{x}^l$ denotes its representation after the activation function of the $l$-th hidden layer~\footnote{The input layer is a special case where $l = 0$, and the output layer is another special case where $l$ is the number of hidden layers plus one. Also note that when $\boldsymbol{x}^l$ is the output layer, the denominator is the gradient of $f$ with respect to $f$ and thus equal to $1$, and equation \eqref{eq:approx_dist_general} reduces to $yf(\boldsymbol{x})$.},
$d_{f, \boldsymbol{x}^l ,y}$ is positive if $\boldsymbol{x}^l$ is correctly classified and negative otherwise.
%
%
%
To mitigate scaling effects that come from measuring plain distances, we normalize $d$ by the square root of the total variation at each layer's inputs $\sqrt{\nu(\boldsymbol{x}^l)}$ as described in \citet{jiang2018predicting}. To gracefully handle small gradients, we add a smoothing constant $\epsilon = 10^{-6}$ to the denominator. Instead of ignoring misclassifications, we squash all outputs of $d$ to the interval [-$\lambda$, $\lambda$] with $z \mapsto \lambda \tanh(\tfrac{z}{\lambda})$, where $\lambda$ is a tunable constant. All these preprocessing give us the transformed margin distribution:
\begin{equation}
\label{eq:approx_dist_transformed}
    \hat{d}_{f, \boldsymbol{x}^l, y} =
    \lambda \tanh\bigg{(}\frac{yf(\boldsymbol{x}^l)}{\lambda (\|\nabla_{\boldsymbol{x}^l} f(\boldsymbol{x}^l) \|_2 + \epsilon)}\bigg{)}.
\end{equation}

Instead of working directly with the whole distribution, we work with a compact signature consisting of the $\{5, 25, 50, 75, 95\}^{th}$ percentiles.
Each DNN of with $L$ hidden layers thus produces a sequence of 5-dimensional vectors $[\boldsymbol{\theta}_0, \ldots, \boldsymbol{\theta}_{L+1}]$ with $L+2$ elements. This defines the feature extraction function $\boldsymbol{\Psi}(f) = [\boldsymbol{\theta}_0, \ldots, \boldsymbol{\theta}_{L+1}]$.

\subsection{Modeling variable-depth neural networks}%
\label{sec:features}

Once we obtain the sequence of vectors $[\boldsymbol{\theta}_0, \ldots, \boldsymbol{\theta}_{L+1}]$ for each DNN, the next step is learn the function $\hat{g}$ from them. Since the number of vectors in the sequence varies with the depth of the DNN, we need a way to handle variable-length inputs.

Our first method of handling variable-length inputs is simply to sum over the sequence:
\begin{equation*}
    \Sigma([\boldsymbol{\theta}_0, \ldots, \boldsymbol{\theta}_{L+1}]) = \sum_{i=0}^{L+1}\boldsymbol{\theta}_i = \boldsymbol{\hat{\theta}},
\end{equation*}
where $\Sigma$ reduces the sequence of vectors into a single vector $\boldsymbol{\hat{\theta}}$ with 5 elements. We then use this row vector as input to our linear models and DNNs.

Our other method of handling variable-length inputs is to use recurrent neural networks (or RNNs), and feed in margin signatures for each layer in successive time steps. This allows us to model more complex layer-dependent relationships between the margin signatures and generalization gap.
To summarize our three GGPs for a linear ($h_l$), DNN ($h_d$) and RNN ($h_r$) models:
\begin{equation*}
    \hat{g}_{\,\scaleto{\text{linear}}{4pt}} = \boldsymbol{\Psi} \circ \Sigma \circ h_{l},\quad
    \hat{g}_{\,\scaleto{\text{DNN}}{4pt}} = \boldsymbol{\Psi} \circ \Sigma \circ h_{d},\quad
    \hat{g}_{\,\scaleto{\text{RNN}}{4pt}} = \boldsymbol{\Psi} \circ h_{r}.
\end{equation*}



\subsection{Evaluation}

To quantify the ``closeness'' between $\hat{g}$ and $g$, we compute two evaluation metrics: the mean $L_1$ loss, and the \emph{coefficient of determination} $(R^2)$:
\[
R^2 \triangleq 1 - \frac{\sum_i (\hat{g}(f_i) - g(f_i))^2}{\sum_i (g(f_i) - \frac{1}{n}\sum_j g(f_j))^2},
\]where $i$ and $j$ index the DNN in the test fold. We perform a 5-fold cross validation, measuring the $R^2$ and $L_1$ loss by comparing test-fold predictions against test-fold labels. The closer $R^2$ is to 1 and $L_1$ loss is to 0, the better the fit.

\section{Experiments}
\label{sec:expt}


\begin{figure*}[t]
    \vspace{-0.05in}
    \centerline{\includegraphics[width=0.94\textwidth]{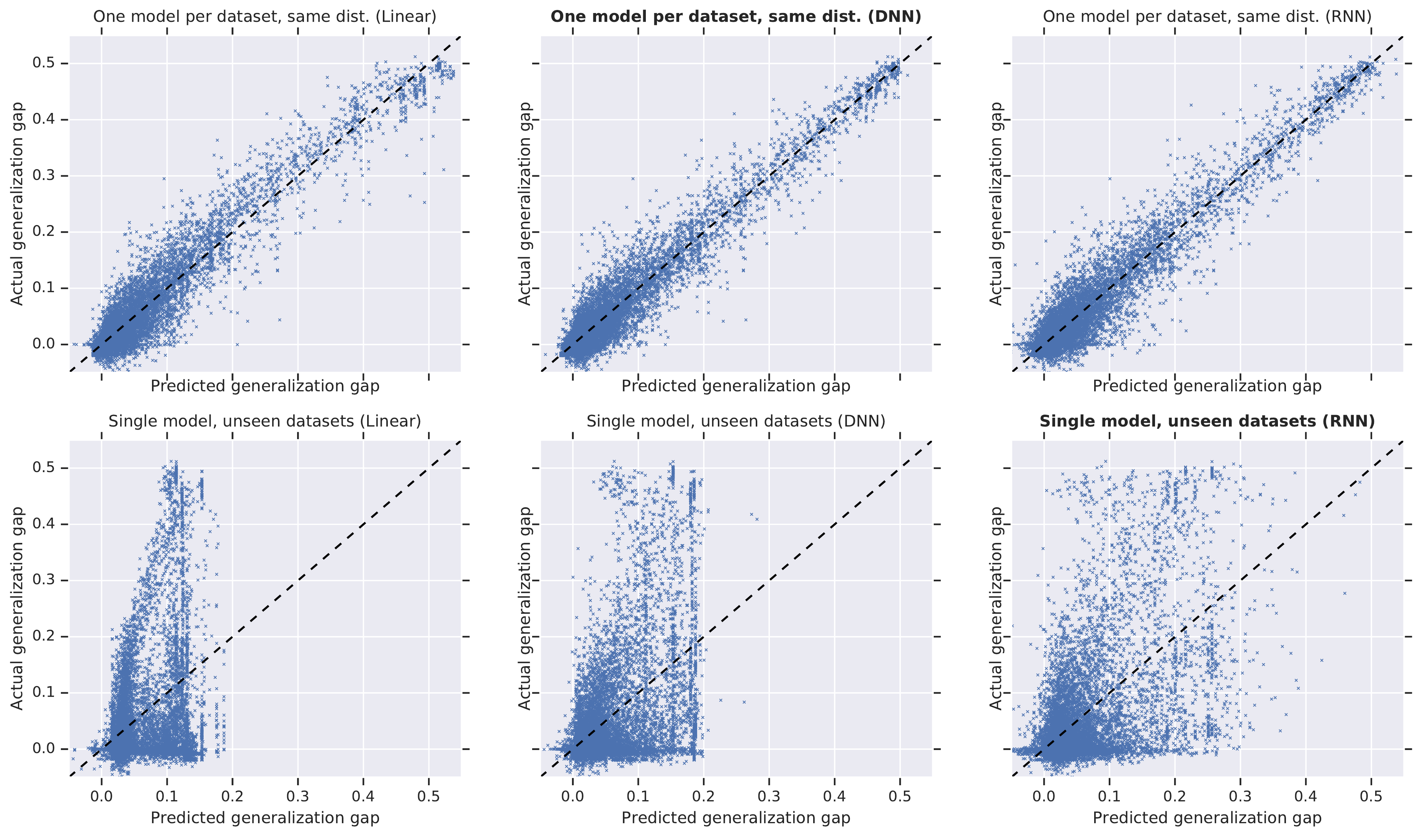}}
    \vspace{-0.1in}
    \caption{Calibration plots for dataset-dependent (top) and dataset-independent (bottom) GGPs. 
    }
    \label{fig:ggp}
\end{figure*}

\begin{table*}[t]
\vskip 0.15in
\begin{center}
\begin{small}
\begin{sc}
\begin{tabular}{c|c|c|c|c|c|c|c|c|c|c}
\toprule
\multirow{3}{*}{Model type} &
\multicolumn{4}{c|}{One model per dataset} &
\multicolumn{6}{c}{Single model} \\
\cmidrule{2-11} &
\multicolumn{2}{c|}{Same dist.} &
\multicolumn{2}{c|}{Unseen hparams} &
\multicolumn{2}{c|}{Same dist.} &
\multicolumn{2}{c|}{Unseen hparams} &
\multicolumn{2}{c}{Unseen datasets} \\
\cmidrule{2-11}
& $R^2$ & $L_1$ loss & $R^2$ & $L_1$ loss & $R^2$ & $L_1$ loss & $R^2$ & $L_1$ loss & $R^2$ & $L_1$ loss \\
\toprule
Linear & 0.956 & 0.0204 & 0.956 & 0.0205 & 0.390 & 0.0658 & 0.384 & 0.0663 & 0.338 & 0.0656 \\
DNN & \textbf{0.965} & \textbf{0.0183} & \textbf{0.964} & \textbf{0.0184} & 0.502 & 0.0592 & 0.487 & \textbf{0.0599} & 0.464 & 0.0595 \\
RNN & 0.959 & 0.0200 & 0.945 & 0.0222 & \textbf{0.584} & \textbf{0.0575} & \textbf{0.527} & 0.0600 & \textbf{0.555} & \textbf{0.0556} \\
\bottomrule
\end{tabular}
\end{sc}
\end{small}
\end{center}
\vskip -0.1in
\caption{Predicting generalization gap with margin signatures.}
\label{tbl:ggp}
\vskip -0.1in
\end{table*}

The goal of the experiments in this section is to investigate the viability of models that can predict the generalization gap of DNNs across multiple datasets and DNN hyperparameters. Additional details are provided in Appendix A.

We generated 27 variations of the spiral dataset~\cite{spiral} that vary by number of loops, amount of input noise, and number of training examples. The data points in each variation are regenerated with 5 random seeds, giving us 135 spiral datasets. We also define 100 DNN hyperparameters that vary by the number of layers, number of hidden units (chosen independently for each layer), optimizer, learning rate, batch size, dropout, and batch normalization. Each of these hyperparameters define a DNN trained for $10^6$ steps on each of the 135 datasets, giving us a total of up to 13,500 trained DNNs (note that DNNs whose losses diverge beyond float32 limits were discarded). The test accuracy of each trained DNN is computed using a test set of size~$10^6$.
We refer to Table \ref{tbl:ggp} for figures on GGPs. All figures are averaged over 5-fold cross validation.

\subsection{Dataset-dependent GGPs}

We first consider the dataset-dependent case -- we group the trained neural networks by the dataset they were trained on, and train one GGP per dataset. We then concatenate over all datasets the predictions and labels, giving us one prediction array and one label array, and compute the $R^2$ and $L_1$ loss from these two arrays.

\textbf{Same distribution}\\
In the \emph{same distribution} task, we split the 13,500 neural networks by the 27 distinct spiral dataset the DNNs were trained on, and obtain 27 disjoint sets of up to 500 DNNs. Within each of these 27 sets, we further split it into 5 folds, \emph{one for each random seed} that was used by the pseudo-random number generator used to generate the spiral dataset.
Note that this is the most similar task to \citet{jiang2018predicting}.

Our results indicate that we are able to replicate the high $R^2$ obtained by \citet{jiang2018predicting} -- our linear model obtained 5-fold cross validation $R^2=0.956$ for the GGP, which is comparable to the $R^2=0.96$ obtained by their best model. Our linear model also obtained $R^2=0.963$ for the test accuracy predictor.

We also observe that both our DNNs and RNNs manage to consistently outperform the linear model, with DNNs performing better -- obtaining an $R^2$ of $0.965$. It's worth noting that all these models are only trained with $\leq 400$ training examples, which is surprising because neural networks, especially RNNs, usually only outperform linear models when trained on large datasets.


\textbf{Extrapolating to unseen hyperparameters}\\
In this task, we similarly group the neural networks by the spiral dataset variations they were trained on, but instead of splitting each group into folds by the random seed, we split them by the hyperparameters that define the neural networks' architecture and optimization, making sure that the training set and test set never contain neural networks with identical hyperparameters. Note that the test set no longer has the same data distribution as the training set.

Surprisingly, most of our GGPs only take a slight penalty in this case -- the $R^2$ of linear and DNN models dropped by no more than $0.002$. This suggests that the GGPs are not merely learning hyperparameter-specific relationships between margin signatures and generalization gap, but are learning relationships that hold more broadly.

\subsection{Dataset-independent GGPs}

Our next set of experiments attempt to make the GGPs \emph{dataset independent}. This task is significantly more challenging than the data-dependent case, since neural networks trained with loosely coiled spirals would tend to have larger input-layer margins, while neural networks trained with tightly coiled spirals would tend to have smaller input-layer margins.
Fortunately, since we combine all the datasets, the models can train on a larger dataset and learn to ignore dataset-specific patterns.

\textbf{Same distribution}\\
We use the same method of splitting the neural networks into 5-folds according to their random seeds, training a single model with 4 folds (around 8000 training examples) and evaluating on the remaining fold. Unsurprisingly, linear models perform poorly on generalization gap prediction ($R^2=0.390$) in this setting, as it now has a harder problem to solve without the expressivity to exploit the extra training examples. With DNNs, we outperform the linear model with $R^2=0.502$, but is still significantly lower than the dataset-dependent case. RNNs ($R^2=0.584$) consistently outperform both linear models and DNNs.

\textbf{Extrapolating to unseen hyperparameters and datasets}\\
We also extrapolate single model GGP to unseen hyperparameters and unseen datasets, and find that both the DNN and RNN models perform only slightly worse than the same-distribution GGP.

\section{Discussion}
We have demonstrated the viability of task-independent GGPs by presenting a new approach to handle aggregated margin features from DNNs.
%
%
We replicated the strong baselines presented by \citet{jiang2018predicting} with a linear model on margin signatures, then established that DNNs and RNNs outperform linear models on almost every generalization gap prediction task. We hope that demonstrating that RNNs can model variable-depth neural network behavior opens up the possibility of modeling more complex neural network architectures, which are directed acyclic graphs in general. We also hope these empirical advancements in modeling when neural networks generalize can serve as a stepping stone towards improved theoretical understanding.

\clearpage


\section*{Acknowledgements}

We thank Yiding Jiang and Dilip Krishnan for sharing their code, data, and answering our questions regarding \citet{jiang2018predicting}. We also thank Charles Weill for the suggestion to use Apache Beam for scaling up our experiments.

\bibliographystyle{icml2019}
\bibliography{main}

\clearpage

\section*{A. Experimental Details}
\label{sec:expt_details}

\subsection*{Generating Spiral Datasets}

The spiral dataset~\cite{spiral} consists of a set of blue and red points on an $x-y$ plane with $x \in [-1, 1]$ and $y \in [-1, 1]$. Blue points lie along the Archimedean spirals that satisfy the polar equation $r = \tfrac{k \theta}{2 \pi}$, while red points lie along $r = \tfrac{k \theta + \pi}{2 \pi}$, where $k$ is the number of loops of the spiral. The binary classification task is: given a point in the $x-y$ plane, is it blue or red?

We vary the ``difficulty'' of each binary classification task in three different ways:
\begin{enumerate}
    \item Number of training examples $m \in \{50, 100, 200\}$.
    \item Number of spiral loops $k \in \{1, 2, 3\}$.
    \item Standard deviation $\sigma$ of the Gaussian noise added to each point, where $\sigma \in \{0, 0.05, 0.15\}.$
\end{enumerate}

All these points are generated from pseudo-random number generators provided by the open source Python library \texttt{numpy}, and for each dataset we regenerate 5 replicas with seeds $\{1, 2, 3, 4, 5\}$. Thus, the set of spiral dataset specifications we generate is simply the Cartesian product of the above sets, which has a size of 135. By storing the specifications by which we generate the data, it allows us to evaluated against an arbitrarily large generated test set without storing the generated data.

\subsection*{Generating DNN Hyperparameters}

We generate 100 sets of DNN hyperparameters, which correspond to the parameters of TensorFlow library's \texttt{tf.estimator.DNNClassifier}. We vary the DNNs in the following ways:
\begin{enumerate}
    \item Number of hidden layers $\in \{1, 2, 3, 4\}$.
    \item Width of each hidden layer $\in \{4, 8, 16\}$. Note that each layer's width varies independently, so it is possible to have a DNN where the first layer has width $4$, the second layer has width $16$, and the third layer has width $4$.
    \item Optimizer $\in \{\text{SGD}, \text{Adam}\}$.
    \item Learning rate $\in \{0.1, 0.01, 0.001\}$.
    \item Batch size $\in \{32, 64, 128\}$. Note that the batch size is always at most the size of the dataset, so when the dataset contains only $50$ training examples, the batch size can only be $32$; when the dataset contains only $100$ training examples, the batch size can be $32$ or $64$, and when the dataset contains $200$ training examples, the batch size can be $32, 64$, or $128$.
    \item Batch norm $\in \{\text{True}, \text{False}\}$.
    \item Dropout $\in \{0, .25, 0.5\}$.
\end{enumerate}
All hidden layers use ReLU activation and Glorot uniform initializer for the kernels and zero as the bias initializer. This gives us more than 10000 possible DNN hyperparameters to choose from. We sample 100 hyperparameters using a fixed random seed, and use them to generate the DNNs that we will later train and analyze.

\subsection*{Training the GGPs}

In this subsection, we describe how each GGP is trained. Since we have to generate 13,500 DNNs, and each DNN takes $\sim 45$ minutes to train and evaluate, we parallelize the training and evaluation tasks using Apache Beam, which reduced the running time from a hypothetical 38 days to 4 hours.

\subsubsection*{Linear}

We use the open source Python package \texttt{sklearn}'s linear\_model.LinearRegression class to perform an ordinary least squares regression. For both dataset-dependent and dataset-independent tasks, we set $\lambda = 0.5$, which squashes all inputs into the range $[-0.5, 0.5]$.

\subsubsection*{DNN}

For the DNN GGP, we use TensorFlow's Keras Layers. We use a DNN with 3 hidden layers, all with 16 hidden units. All hidden layers use ReLU activation and Glorot uniform initializer for the kernels and zero as the bias initializer. We use the Adagrad optimizer with learning rate 0.1, and mean squared error as the loss function. For the dataset-dependent tasks, we set $\lambda = 0.5$ and train for 5000 steps with batch size 64. For the dataset-independent tasks, we set $\lambda = 2.5$ and train for 25000 steps with batch size 64.

\subsubsection*{RNN}

For the RNN GGP, we also use TensorFlow's Keras Layers. We use a RNN where the first layer is a \texttt{SimpleRNN} layer with 16 hidden units and $\tanh$ activation. This layer reduces over the number of rows in the feature matrix $\boldsymbol{\theta}$, and outputs a $16$-dimensional vector for each feature matrix. This vector is then passed through 3 more hidden layers, each with 16 hidden units and ReLU activation. We use the Adagrad optimizer with learning rate 0.1, and mean squared error as the loss function. For the dataset-dependent tasks, we set $\lambda = 0.5$ and train for 2500 steps with batch size 64. For the dataset-independent tasks, we set $\lambda = 2.5$ and train for 25000 steps with batch size 64.

\clearpage

\onecolumn
\section*{B. Predicting Test Accuracy from Margin Signatures}

\begin{figure}[H]
    \vskip -0.1in
    \centerline{\includegraphics[width=0.94\textwidth]{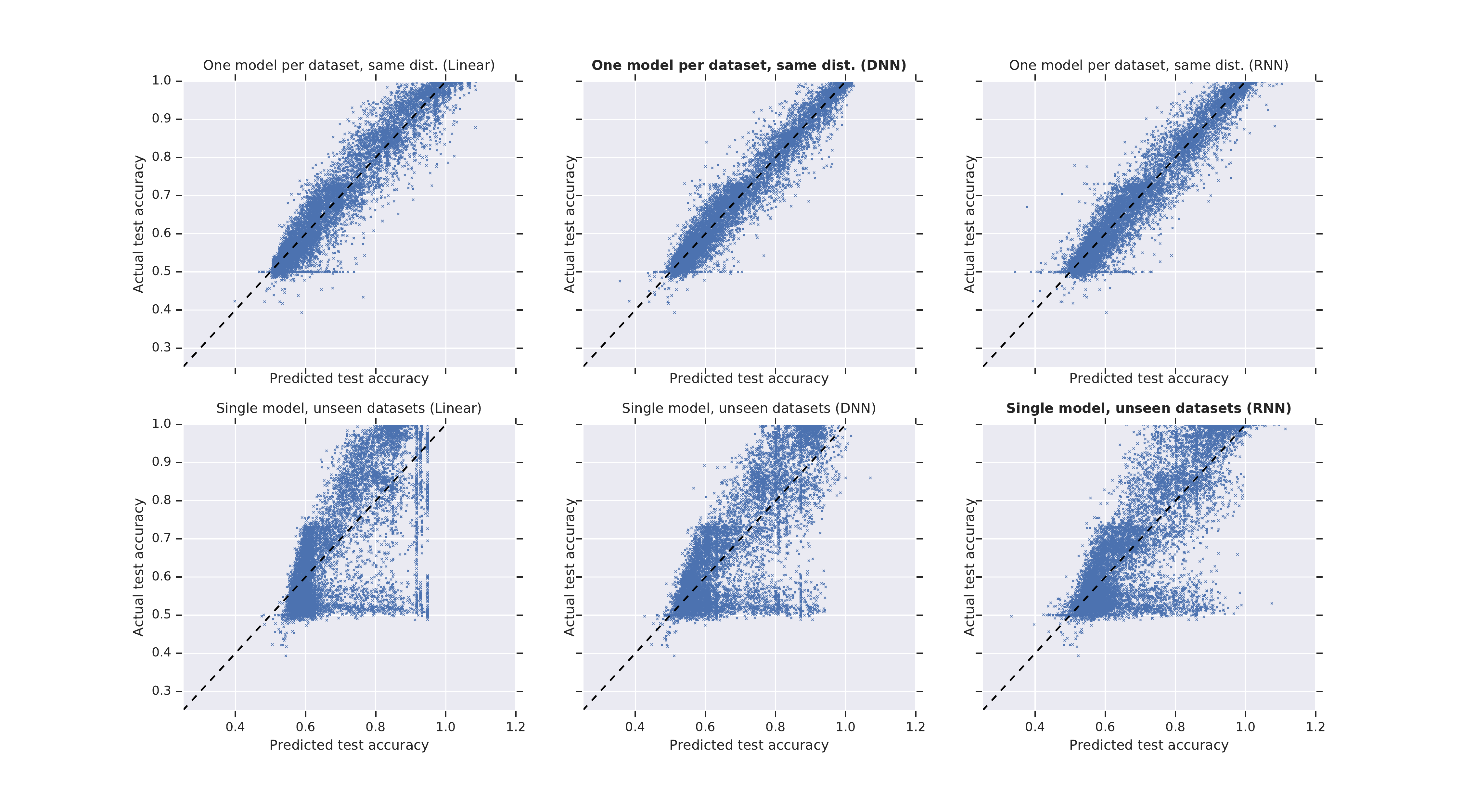}}
    \vskip -0.3in
    \caption{Calibration plots for dataset-dependent (top) and dataset-independent (bottom) test accuracy predictors.}
    \label{fig:test_acc}
\end{figure}
\begin{table}[H]
\vskip -0.15in
\begin{center}
\begin{small}
\begin{sc}
\begin{tabular}{c|c|c|c|c|c|c|c|c|c|c}
\toprule
\multirow{3}{*}{Model type} &
\multicolumn{4}{c|}{One model per dataset} &
\multicolumn{6}{c}{Single model} \\
\cmidrule{2-11} &
\multicolumn{2}{c|}{Same dist.} &
\multicolumn{2}{c|}{Unseen hparams} &
\multicolumn{2}{c|}{Same dist.} &
\multicolumn{2}{c|}{Unseen hparams} &
\multicolumn{2}{c}{Unseen datasets} \\
\cmidrule{2-11}
& $R^2$ & $L_1$ loss & $R^2$ & $L_1$ loss & $R^2$ & $L_1$ loss & $R^2$ & $L_1$ loss & $R^2$ & $L_1$ loss \\
\toprule
Linear & 0.963 & 0.0309 & 0.961 & 0.0316 & 0.736 & 0.0835 & 0.730 & 0.0837 & 0.709 & 0.0834 \\
DNN & \textbf{0.978} & \textbf{0.0236} & \textbf{0.977} & \textbf{0.0238} & 0.808 & 0.0699 & 0.802 & 0.0700 & 0.781 & 0.0705 \\
RNN & 0.971 & 0.0272 & 0.961 & 0.0307 & \textbf{0.834} & \textbf{0.0636} & \textbf{0.815} & \textbf{0.0664} & \textbf{0.802} & \textbf{0.0663} \\
\bottomrule
\end{tabular}
\end{sc}
\end{small}
\caption{Predicting test accuracy from margin signatures.}
\label{tbl:test_acc}
\end{center}
\vskip -0.1in
\end{table}

We repeated the same experimental setups in section \ref{sec:expt}, but using \emph{test accuracy} instead of generalization gap as the label for the regression problem. The above figures show our results. Observe that many patterns that hold for GGPs also hold for test accuracy predictors:
\begin{itemize}
    \item Linear models already obtain a very high coefficient of determination ($R^2=0.963$), but RNNs and DNNs both outperform them, with DNNs having the best performance ($R^2=0.978$).
    \item Dataset-independent models uniformly perform worse than dataset-dependent models, with RNNs $>$ DNNs $>$ linear models.
    \item All models perform slightly worse on out-of-distributions tasks than same-distribution tasks.
\end{itemize}
However, a seemingly puzzling observation is that compared to the GGP figures, test accuracy prediction figures uniformly have higher $R^2$ \emph{and} higher $L_1$ loss. So which task is ``easier''? We believe the answer is ``neither''. Recall that $R^2 \triangleq 1 - \tfrac{\text{residuals}}{\text{variance}}$. This means that both $R^2$ and $L_1$ loss can be higher if test accuracy predictors have larger residuals relative to GGPs, but the test accuracy labels have even larger variance. Thus, while test accuracy predictors look better than GGPs in terms of $R^2$, if we are directly comparing which model approximates the validation set more closely, we ought to be comparing the $L_1$ loss. However, if we are allowed to tune a scaling factor and intercept for the predictor, then we ought to be comparing the $R^2$ (but we think this scenario makes these models less practical).
\clearpage

\section*{C. Analysis of the training data}
The following section contains an analysis of the data used to train the generalization gap predictor. These data is the results of the aggregation of the margin features described in Section~\ref{sec:features}. All figures describe the relation between train accuracy and generalization gap. As it can be seen, high dropout and high learning rate correlate with lower accuracy and low generalization gap prediction. However, there doesn't seem to be a clear direct correlation between batch normalization, batch size and generalization gap.
\begin{figure}[ht] 
  \noindent\begin{minipage}[t]{0.48\linewidth}
    \centering
    \includegraphics[width=\linewidth]{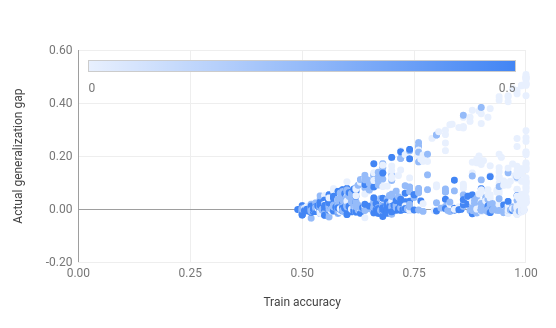}
    \caption{Generalization gap by train accuracy, dropout indicates the magnitude of the points.}
    \vspace{0.3in}
  \end{minipage}\hspace{0.3in}%
  \begin{minipage}[t]{0.48\linewidth}
    \centering
    \includegraphics[width=\linewidth]{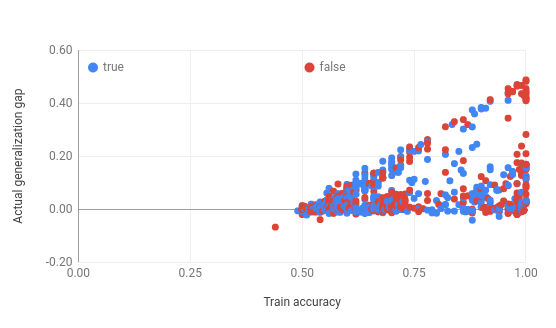}
    \caption{Generalization gap by train accuracy, batch normalization is red for activated and blue for deactivated.}
    \vspace{0.3in}
  \end{minipage}
  \begin{minipage}[t]{0.48\linewidth}
    \centering
    \includegraphics[width=\linewidth]{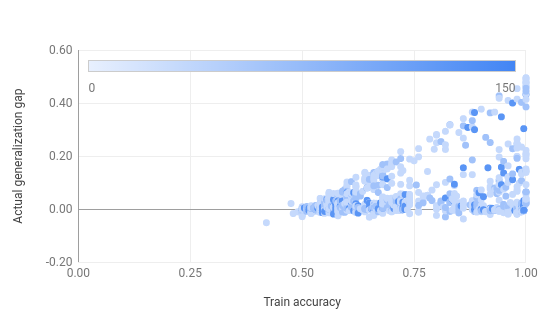}
    \caption{Generalization gap by train accuracy, batch size indicates the magnitude of the points.}
    \vspace{0.3in}
  \end{minipage}\hspace{0.3in}%
  \begin{minipage}[t]{0.48\linewidth}
    \centering
    \includegraphics[width=\linewidth]{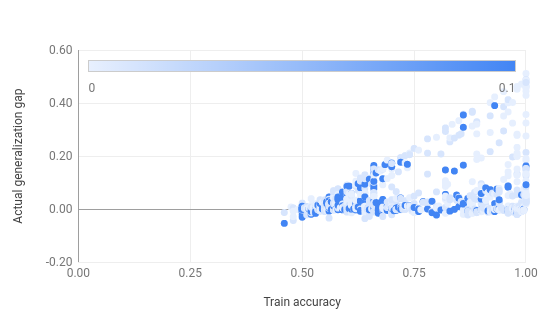}
    \caption{Generalization gap by train accuracy, learning rate indicates the magnitude of the points.}
  \end{minipage}
\end{figure}


\end{document}